\documentclass{article}

\usepackage[preprint,nonatbib]{nips_2018}
\usepackage[square,sort,comma,numbers]{natbib}

\usepackage[utf8]{inputenc} 
\usepackage[T1]{fontenc}    
\usepackage{hyperref}       
\hypersetup{
    colorlinks=true,
    linkcolor=blue,
    filecolor=magenta,      
    urlcolor=blue,
}
\usepackage{url}            
\usepackage{booktabs}       
\usepackage{amsfonts}       
\usepackage{nicefrac}       
\usepackage{microtype}      

\usepackage{algorithm}
\usepackage{algorithmic}
\usepackage{wrapfig}
\usepackage{enumitem}
\usepackage{subfigure}

\usepackage{amssymb,amsmath}
\usepackage{mathtools}

\title{Reinforcement Learning for Improving Agent Design}

\author{
  David Ha \\
  Google Brain\\
  Tokyo, Japan \\
  \texttt{hadavid@google.com} \\
}

\begin{document}

\maketitle

\begin{abstract}
In many reinforcement learning tasks, the goal is to learn a policy to manipulate an agent, whose design is fixed, to maximize some notion of cumulative reward. The design of the agent's physical structure is rarely optimized for the task at hand. In this work, we explore the possibility of learning a version of the agent's design that is better suited for its task, jointly with the policy. We propose an alteration to the popular OpenAI Gym framework, where we parameterize parts of an environment, and allow an agent to jointly learn to modify these environment parameters along with its policy. We demonstrate that an agent can learn a better structure of its body that is not only better suited for the task, but also facilitates policy learning. Joint learning of policy and structure may even uncover design principles that are useful for assisted-design applications. Videos of results at \url{https://designrl.github.io/}.
\end{abstract}

\section{Introduction}

Embodied cognition~\cite{anderson2003embodied,mahon2008critical,shapiro2010embodied} is the theory that an organism's cognitive abilities is shaped by its body. It is even argued that an agent's cognition extends beyond its brain, and is strongly influenced by aspects of its body and also the experiences from its various sensorimotor functions~\cite{wilson2002six,gover1996embodied}. Evolution plays a vital role in shaping an organism's body to adapt to its environment; the brain and its ability to learn is only one of many body components that is co-evolved together~\cite{pfeifer2006body,bongard2011morphological}. We can recognize embodiment in nature by observing that many organisms exhibit complex motor skills, such as the ability to jump~\cite{bresadola1998medicine} or swim~\cite{beal2006passive} even after brain death.

\begin{figure}[!htb]
\vskip -0.05in
\begin{center}
\centerline{\includegraphics[width=1.0\columnwidth]{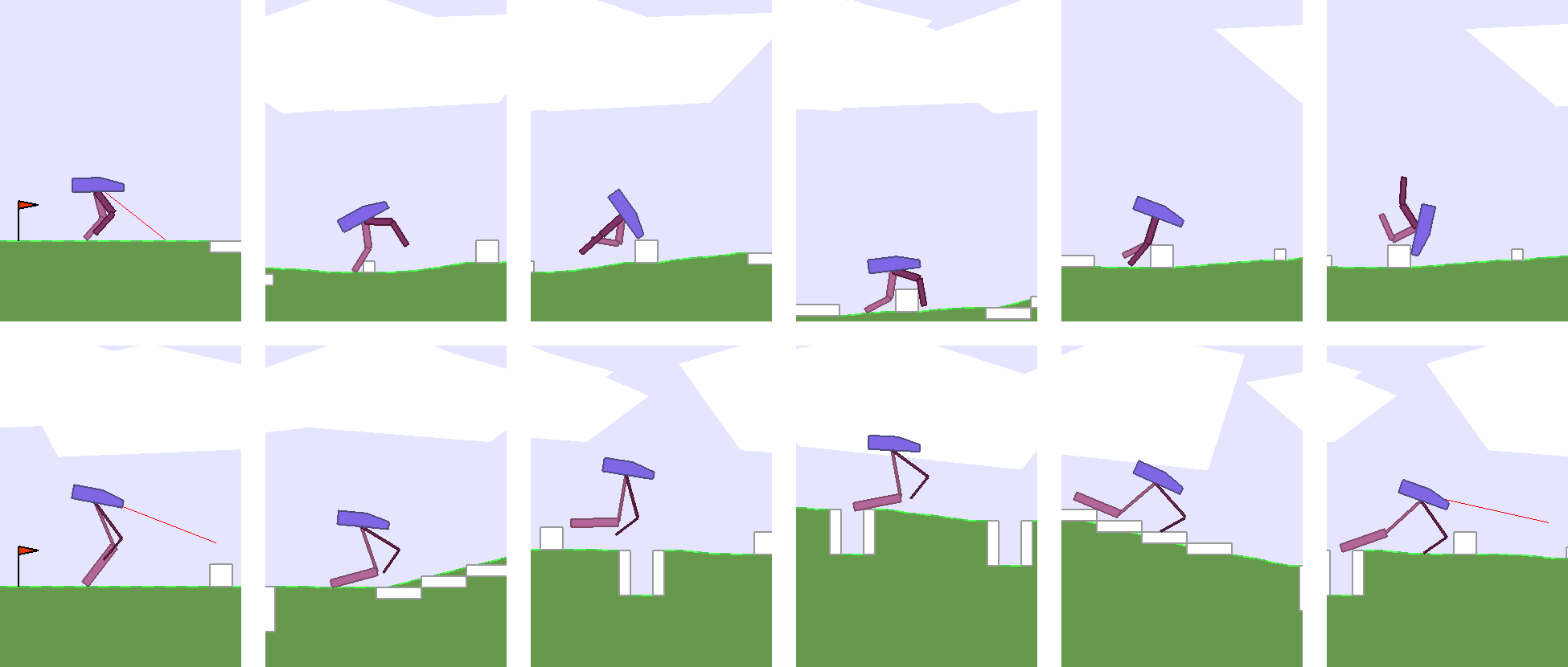}}
\vskip -0.07 in
\caption{Learning to navigate over randomly generated terrain in \texttt{BipedalWalkerHardcore-v2} environment~\cite{openai_gym} (top). Agent learns a better body design while jointly learning to navigate (bottom).}
\label{fig:bipedhard}
\end{center}
\vskip -0.20 in
\end{figure}

While evolution shapes the overall structure of the body of a particular species, an organism can also change and adapt its body to its environment during its life. For instance, professional athletes spend their lives body training while also improving specific mental skills required to master a particular sport~\cite{tricoli2005short}. In everyday life, regular exercise not only strengthens the body but also improves mental conditions~\cite{raglin1990exercise,deslandes2009exercise}. We not only learn and improve our skills and abilities during our lives, but also learn to shape our bodies for the lives we want to live.

We are interested to investigate embodied cognition within the reinforcement learning (RL) framework. Most baseline tasks~\cite{todorov2012mujoco,roboschool} in the RL literature test an algorithm's ability to learn a policy to control the actions of an agent, with a predetermined body design, to accomplish a given task inside an environment. The design of the agent's body is rarely optimal for the task, and sometimes even intentionally designed to make policy search challenging. In this work, we explore enabling learning versions of an agent's body that are better suited for its task, jointly with its policy. We demonstrate that an agent can learn a better structure of its body that is not only better for its task, but also facilitates policy learning. We can even optimize our agent's body for certain desired characteristics, such as material usage. Our approach may help uncover design principles useful for assisted-design.

Furthermore, we believe the ability to learn useful morphology is an important area for the advancement of AI. Although morphology learning originally initiated from the field of evolutionary computation, there has also been great advances in RL in recent years, and we believe much of what happens in ALife should be in principle be of interest to the RL community and vice versa, since learning and evolution are just two sides of the same coin.

We believe that conducting experiments using standardized simulation environments facilitate the communication of ideas across disciplines, and for this reason we design our experiments based on applying ideas from ALife, namely morphology learning, to standardized tasks in the OpenAI Gym environment, a popular testbed for conducting experiments in the RL community. We decide to use standardized Gym environments such as Ant (based on Bullet physics engine) and Bipedal Walker (based on Box2D) not only for their simplicity, but also because their difficulty is well-understood due to the large number of RL publications that use them as benchmarks. As we shall see later, the \texttt{BipedalWalkerHardcore-v2} task, while simple looking, is especially difficult to solve with modern Deep RL methods. By applying simple morphology learning concepts from ALife, we are able to make a difficult task solvable with much fewer compute resources. We also made the code for augmenting OpenAI Gym for morphology learning, along with all pretrained models for reproducing results in this paper available at \url{https://github.com/hardmaru/astool}.

We hope this paper can serve as a catalyst to precipitate a cultural shift in both fields and encourage researchers to open up our minds to each other. By drawing ideas from ALife and demonstrating them in the OpenAI Gym platform used by RL, we hope this work can set an example to bring both the RL and ALife communities closer together to find synergies and push the AI field forward.

\section{Related Work}
There is a broad literature in evolutionary computation, artificial life and robotics devoted to studying, and modelling embodied cognition~\cite{pfeifer2006body}. In 1994, Karl Sims demonstrated that artificial evolution can produce novel morphology that resemble organisms observed in nature~\cite{sims1994evolving,sims1994evolving_MIT}. Subsequent works further investigated morphology evolution~\cite{bongard2011morphological,auerbach2012relationship,auerbach2014environmental,leger1999automated,szerlip2013indirectly,szerlip2014steps,moore2014evolutionary,boxcar2d,auerbach2014robogen}, modular robotics~\cite{lipson2000automatic,ostergaard2003evolving,prokopenko2006evolving,zykov2007evolved}, and evolving soft robots~\cite{cheney2013unshackling,corucci2018evolving}, using indirect encoding~\cite{neat,hyperneat,auerbach2010evolving,auerbach2011evolving,auerbach2010dynamic}.

Literature in the area of passive dynamics study robot designs that rely on natural swings of motion of body components instead of deploying and controlling motors at each joint~\cite{mcgeer1990passive,collins2001three,paul2004morphology,collins2005efficient}. Notably, artist Theo Jansen~\cite{jansen2008strandbeests} also employed evolutionary computation to design physical \textit{Strandbeests} that can walk on their own consuming only wind energy to raise environmental awareness.

Recent works in robotics investigate simultaneously optimizing body design and control of a legged robot~\cite{ha2017joint,ha2018computational} using constraint-based modelling, which is related to our RL-based approach. Related to our work, \cite{geijtenbeek2013flexible,agrawal2014diverse} employ CMA-ES~\cite{cmaes} to optimize over both the motion control and physical configuration of agents. A related recent work~\cite{schaff2018jointly,schaff2018jointly_iclr_workshop} employs RL to learn both the policy and design parameters in an alternating fashion, where a single shared policy controls a distribution of different designs, while in this work we simply treat both policy and design parameters the same way.

\section{Method}

In this section, we describe the method used for learning a version of the agent's design better suited for its task jointly with its policy. In addition to the weight parameters of our agent's policy network, we will also parameterize the agent's environment, which includes the specification of the agent's body structure. This extra parameter vector, which may govern the properties of items such as width, length, radius, mass, and orientation of an agent's body parts and their joints, will also be treated as a learnable parameter. Hence the weights $w$ we need to learn will be the parameters of the agent's policy network combined with the environment's parameterization vector. During a rollout, an agent initialized with $w$ will be deployed in an environment that is also parameterized  with the same parameter vector $w$.

\begin{figure}[!htb]
\vskip -0.01in
\begin{center}
\centerline{\includegraphics[width=1.0\columnwidth]{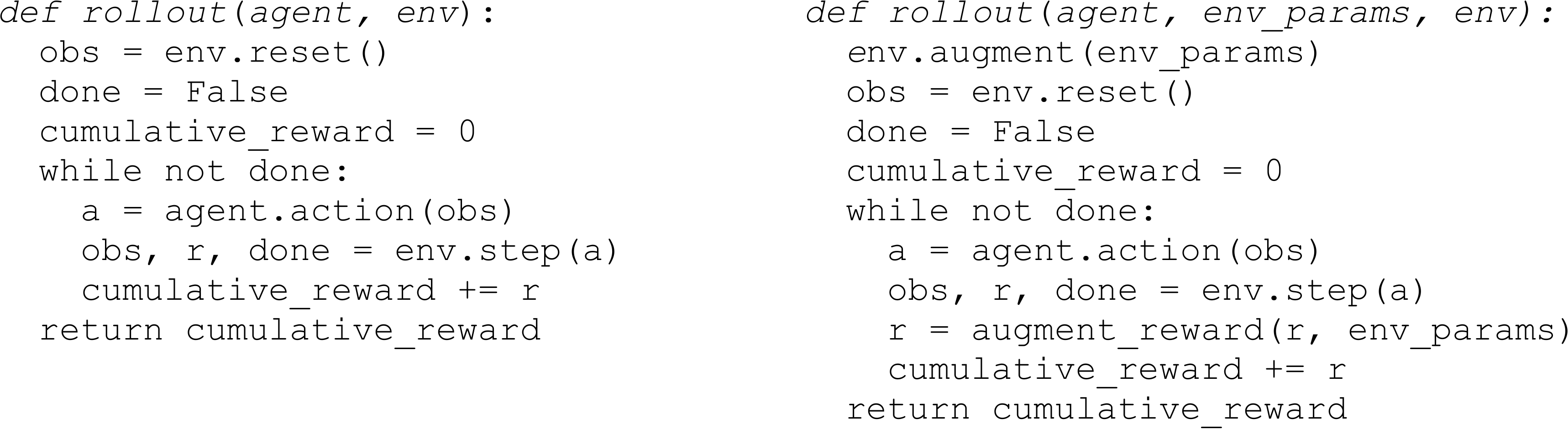}}
\vskip -0.05 in
\caption{OpenAI Gym~\cite{openai_gym} framework for rolling out an agent in an environment (left). We propose an alteration where we parameterize parts of an environment, and allow an agent to modify its environment before a rollout, and also augment its reward based on these parameters (right).}
\label{fig:openai_gym_augment}
\end{center}
\vskip -0.10 in
\end{figure}

The goal is to learn $w$ to maximize the expected cumulative reward, $E[R(w)]$, of an agent acting on a policy with parameters $w$ in an environment governed by the same $w$. In our approach, we search for $w$ using a population-based policy gradient method based on Section 6 of Williams' 1992 REINFORCE~\cite{williams1992}. The next section provides an overview of this algorithm.

Armed with the ability to change the design configuration of an agent's own body, we also wish to explore encouraging the agent to challenge itself by rewarding it for trying more difficult designs. For instance, carrying the same payload using smaller legs may result in a higher reward than using larger legs. Hence the reward given to the agent may also be augmented according to its parameterized environment vector. We will discuss reward augmentation to optimize for desirable design properties later on in more detail in Section~\ref{section:optimizedesignproperties}. 

\subsection{Overview of Population-based Policy Gradient Method (REINFORCE)}
\label{reinforce}
In this section we provide an overview of the population-based policy gradient method described in Section 6 of William's REINFORCE~\cite{williams1992} paper for learning a parameter vector $w$ in a reinforcement learning environment. In this approach, $w$ is sampled from a probability distribution $\pi(w, \theta)$ parameterized by $\theta$. We define the expected cumulative reward $R$ as:

\vskip -0.00 in
\begin{equation}
J(\theta) = E_{\theta}[R(w)] = \int R(w) \; \pi(w, \theta) \; dw.
\end{equation}
\vskip -0.00 in

Using the \textit{log-likelihood trick} allows us to write the gradient of $J(\theta)$ with respect to $\theta$:

\vskip -0.00 in
\begin{equation}
\nabla_{\theta} J(\theta) = E_{\theta}[ \; R(w)  \; \nabla_{\theta} \log \pi(w, \theta) \; ].
\end{equation}
\vskip -0.00 in

In a population size of $N$, where we have solutions $w^1$, $w^2$, ..., $w^N$, we can estimate this as:

\vskip -0.00 in
\begin{equation}
\label{eq:gradient_approx}
\nabla_{\theta} J(\theta) \approx \frac{1}{N} \sum_{i=1}^{N} \; R(w^i)  \; \nabla_{\theta} \log \pi(w^i, \theta).
\end{equation}
\vskip -0.00 in

With this approximated gradient $\nabla_{\theta} J(\theta)$, we then can optimize $\theta$ using gradient ascent:

\vskip -0.00 in
\begin{equation}
\label{eq:theta_update}
\theta \rightarrow \theta + \alpha \nabla_{\theta} J(\theta)
\end{equation}
\vskip -0.00 in

and sample a new set of candidate solutions $w$ from updating the pdf using learning rate $\alpha$. We follow the approach in REINFORCE where $\pi$ is modelled as a factored multi-variate normal distribution. Williams derived closed-form formulas of the gradient $\nabla_{\theta} \log \pi(w^i, \theta)$. In this special case, $\theta$ will be the set of mean $\mu$ and standard deviation $\sigma$ parameters. Therefore, each element of a solution can be sampled from a univariate normal distribution $w_j \sim N(\mu_j, \sigma_j)$. Williams derived the closed-form formulas for the $\nabla_{\theta} \log N(z^i, \theta)$ term in Equation~\ref{eq:gradient_approx}, for each individual $\mu$ and $\sigma$ element of vector $\theta$ on each solution $i$ in the population:

\begin{equation}
\label{eq:mu_sigma_update}
\nabla_{\mu_{j}} \log N(w^i, \theta) = \frac{w_j^i - \mu_j}{\sigma_j^2}, \; 
\nabla_{\sigma_{j}} \log N(w^i, \theta) = \frac{(w_j^i - \mu_j)^2 - \sigma_j^2}{\sigma_j^3}.
\end{equation}

For clarity, we use subscript $j$, to count across parameter space in $w$, and this is not to be confused with superscript $i$, used to count across each sampled member of the population of size $N$. Combining Equations~\ref{eq:mu_sigma_update} with Equation~\ref{eq:theta_update}, we can update $\mu_j$ and $\sigma_j$ at each generation via a gradient update.

We note that there is a connection between population-based REINFORCE, a population-based policy gradient method, and particular formulations of Evolution Strategies~\cite{rechenberg1978evolutionsstrategien,schwefel1981numerical}, namely ones that are not elitist. For instance, Natural Evolution Strategies (NES)~\cite{pepg,wierstra2008natural} and OpenAI-ES~\cite{openai_es} are closely based on Sec. 6 of REINFORCE. There is also a connection between natural gradients (computed using NES) and CMA-ES~\cite{cmaes}. We refer to Akimoto et al.~\cite{akimoto2012theoretical} for a detailed theoretical treatment and discussion about the connection between CMA-ES and natural gradients. 

\section{Experiments}

In this work, we experiment on the continuous control \texttt{RoboschoolAnt-v1} environment from Roboschool~\cite{roboschool}, based on the open source Bullet~\cite{pybullet} physics engine, and also \texttt{BipedalWalker-v2} from the Box2D~\cite{box2d} section of the OpenAI Gym~\cite{openai_gym} set of environments. For simplicity, we first present results of anecdotal examples obtained over a single representative experimental run to convey qualitative results such as morphology and its relationship to performance. A more comprehensive quantitative study based on multiple runs using different random seeds will be presented in Section~\ref{section:multiple_runs}.

\begin{figure}[!htb]
\vskip -0.05in
\begin{center}
\centerline{\includegraphics[width=1.0\columnwidth]{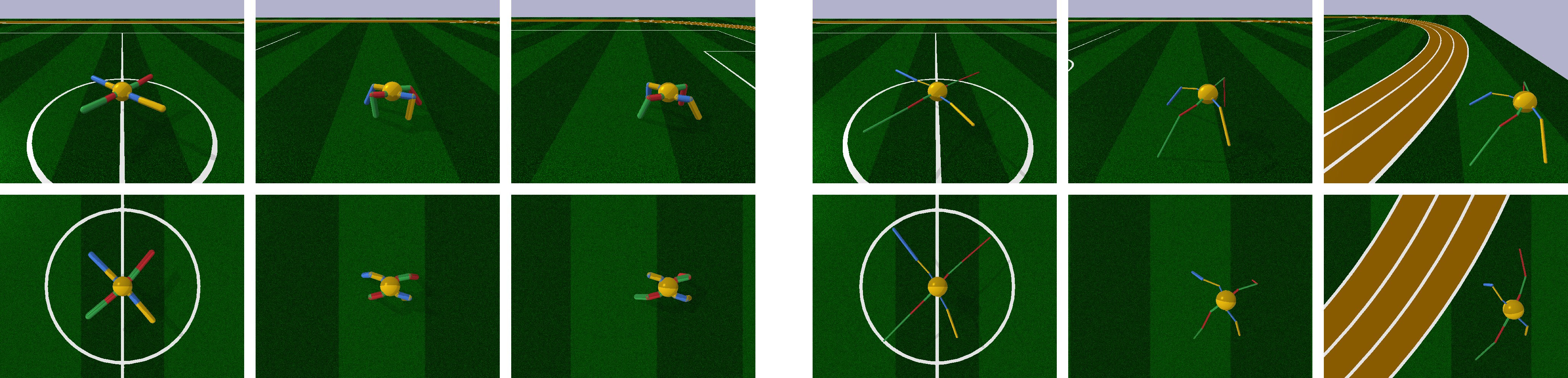}}
\vskip -0.07 in
\caption{Agent learning a policy to navigate forward in \texttt{RoboschoolAnt-v1} environment~\cite{roboschool} (left). Agent develops longer, thinner legs while supporting the same body during training (right).}
\label{fig:bullet_ant}
\end{center}
\vskip -0.20 in
\end{figure}

The \texttt{RoboschoolAnt-v1}\footnote{A compatible version of this environment is also available in PyBullet~\cite{pybullet} which was used for visualization.} environment features a four-legged agent called the \textit{Ant}
. The body is supported by 4 legs, and each leg consists of 3 parts which are controlled by 2 motor joints. The bottom left diagram of Figure~\ref{fig:bullet_ant} describes the initial orientation of the agent. The length of each part of a leg is controlled by the $\Delta x$ and $\Delta y$ distances from its joint connection. A size parameter also controls the radius of each leg part.

In our experiment, we keep the volumetric mass density of all materials, along with the parameters of the motor joints identical to the original environment, and allow the 36 parameters (3 parameters per leg part, 3 leg parts per leg, 4 legs in total) to be learned. In particular, we allow each part to be scaled to a range of $\pm$ 75\% of its original value. This allows us to keep the sign and direction for each part to preserve the original intended structure of the design.

\begin{table}[!htb]
\begin{center}
\begin{small}
\begin{tabular}{lllllllll}
\toprule
 &  \multicolumn{2}{l}{\textbf{Top Left}} &  \multicolumn{2}{l}{\textbf{Top Right}} & \multicolumn{2}{l}{\textbf{Bottom Left}}  &  \multicolumn{2}{l}{\textbf{Bottom Right}}  \\
 & Length & Radius & Length & Radius & Length & Radius & Length & Radius \\
\midrule
Top & 141\% & 33\% & 141\% & 25\% & 169\% & 35\% & 84\% & 51\% \\
Middle & 169\% & 26\% & 164\% & 26\% & 171\% & 31\% & 140\% & 29\% \\
Bottom & 174\% & 26\% & 168\% & 50\% & 173\% & 29\% & 133\% & 38\% \\
\bottomrule
\end{tabular}
\end{small}
\end{center}
\caption{Learned agent body for \texttt{RoboschoolAnt-v1} as \% of original design specification.}
\label{tab:ant}
\vskip -0.20 in
\end{table}

Figure~\ref{fig:bullet_ant} illustrates the learned agent design compared to the original design. With the exception of one leg part, it learns to develop longer, thinner legs while jointly learning to carry the body across the environment. While the original design is symmetric, the learned design (Table~\ref{tab:ant}) breaks symmetry, and biases towards larger rear legs while jointly learning the navigation policy using an asymmetric body. The original agent achieved an average cumulative score of 3447 $\pm$ 251 over 100 trials, compared to 5789 $\pm$ 479 for an agent that learned a better body design.

The Bipedal Walker series of environments is based on the Box2D~\cite{box2d} physics engine. Guided by LIDAR sensors, the agent is required to navigate across an environment of randomly generated terrain within a time limit, without falling over. The agent's payload -- its head, is supported by 2 legs. The top and bottom parts of each leg is controlled by two motor joints. In the easier \texttt{BipedalWalker-v2}~\cite{bipedalwalker} environment, the agent needs to travel across small random variations of a flat terrain. The task is considered solved if an agent obtains an average score greater than 300 points over 100 rollouts.

\begin{figure}[!htb]
\vskip -0.05in
\begin{center}
\centerline{\includegraphics[width=1.0\columnwidth]{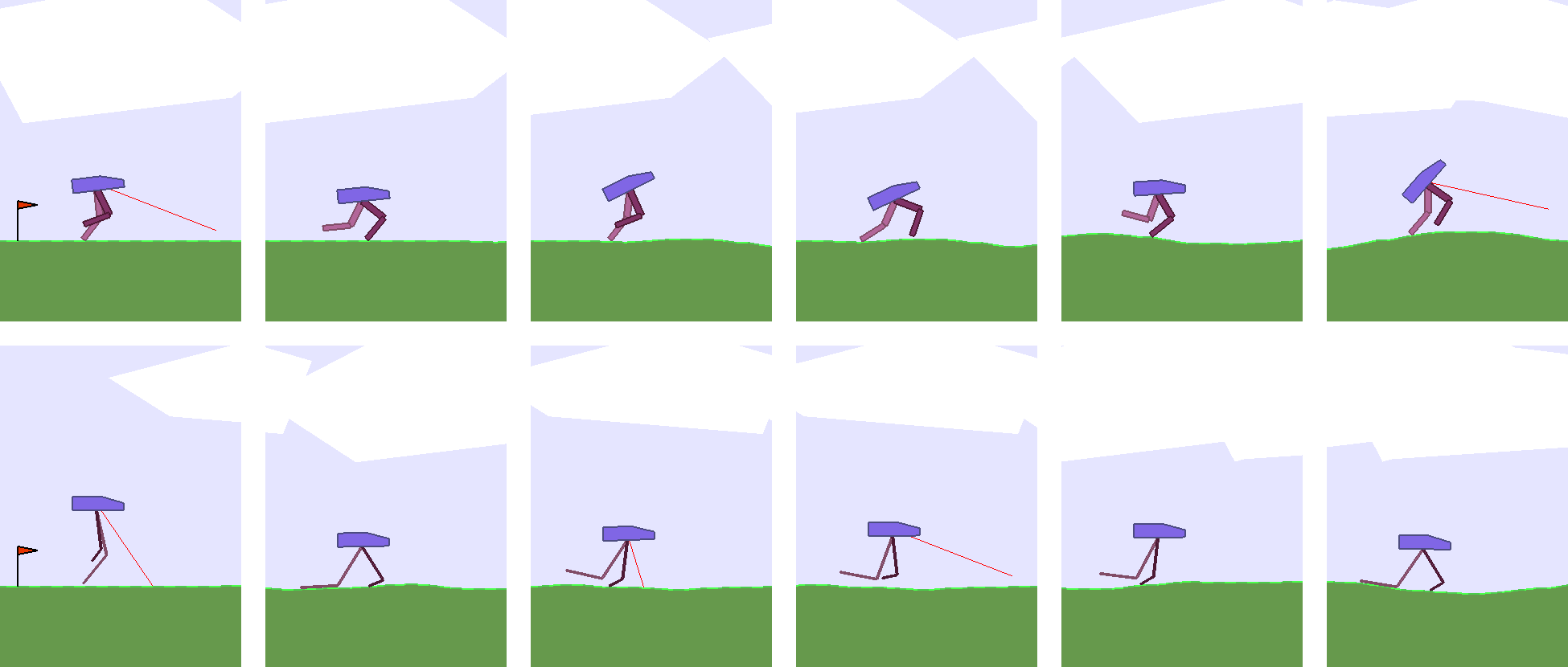}}
\vskip -0.07 in
\caption{Agent learning a policy to navigate forward in \texttt{BipedalWalker-v2} environment~\cite{openai_gym} (top). Agent learns a body to allow it to bounce forward efficiently (bottom).}
\label{fig:biped}
\end{center}
\vskip -0.20 in
\end{figure}

Keeping the head payload constant, and also keeping the density of materials and the configuration of the motor joints the same as the original environment, we only allow the lengths and widths for each of the 4 leg parts to be learnable, subject to the same range limit of $\pm$ 75\% of the original design. In the original environment in Figure~\ref{fig:biped} (top), the agent learns a policy that is reminiscent of a joyful skip across the terrain, achieving an average score of 347. In the learned version in Figure~\ref{fig:biped} (bottom), the agent's policy is to hop across the terrain using its legs as a pair of springs, achieving a score of 359.

In our experiments, all agents were implemented using 3 layer fully-connected networks with $\tanh$ activations. The agent in \texttt{RoboschoolAnt-v1} has 28 inputs and 8 outputs, all bounded between $-1$ and $+1$, with hidden layers of 64 and 32 units. The agents in \texttt{BipedalWalker-v2} and \texttt{BipedalWalkerHardcore-v2} has 24 inputs and 4 outputs all bounded between $-1$ and $+1$, with 2 hidden layers of 40 units each.

Our population-based training experiments were conducted on 96-CPU core machines. Following the approach described in \cite{stablees}, we used a population size of 192, and had each agent perform the task 16 times with different initial random seeds. The agent's reward signal used by the policy gradient method is the average reward of the 16 rollouts. The most challenging BipedalWalkerHardcore agents were trained for 10000 generations, while the easier BipedalWalker and Ant agents were trained for 5000 and 3000 generations respectively. As done in \cite{stablees}, we save the parameters of the agent that achieves the best average cumulative reward during its entire training history.

\subsection{Joint learning of body design facilitates policy learning: \texttt{BipedalWalkerHardcore-v2}}

Learning a better version of an agent's body not only helps achieve better performance, but also enables the agent to jointly learn policies more efficiently. We demonstrate this in the much more challenging \texttt{BipedalWalkerHardcore-v2}~\cite{bipedalwalkerhardcore} version of the task. Unlike the easier version, the agent must also learn to walk over obstacles, travel up and down hilly terrain, and even jump over pits. Figure~\ref{fig:bipedhard} illustrates the original and learnable versions of the environment.\footnote{As of writing, two methods are reported to solve this task. Population-based training~\cite{stablees} (our baseline), solves this task in 40 hours on a 96-CPU machine, using a small feed forward policy network. A3C~\cite{mnih2016asynchronous} adapted for continuous control~\cite{griffis2018} solves the task in 48 hours on a 72-CPU machine, but requires an LSTM~\cite{lstm} policy.}

\begin{figure}[!htb]
\vskip -0.05in
\begin{center}
\centerline{\includegraphics[width=1.0\columnwidth]{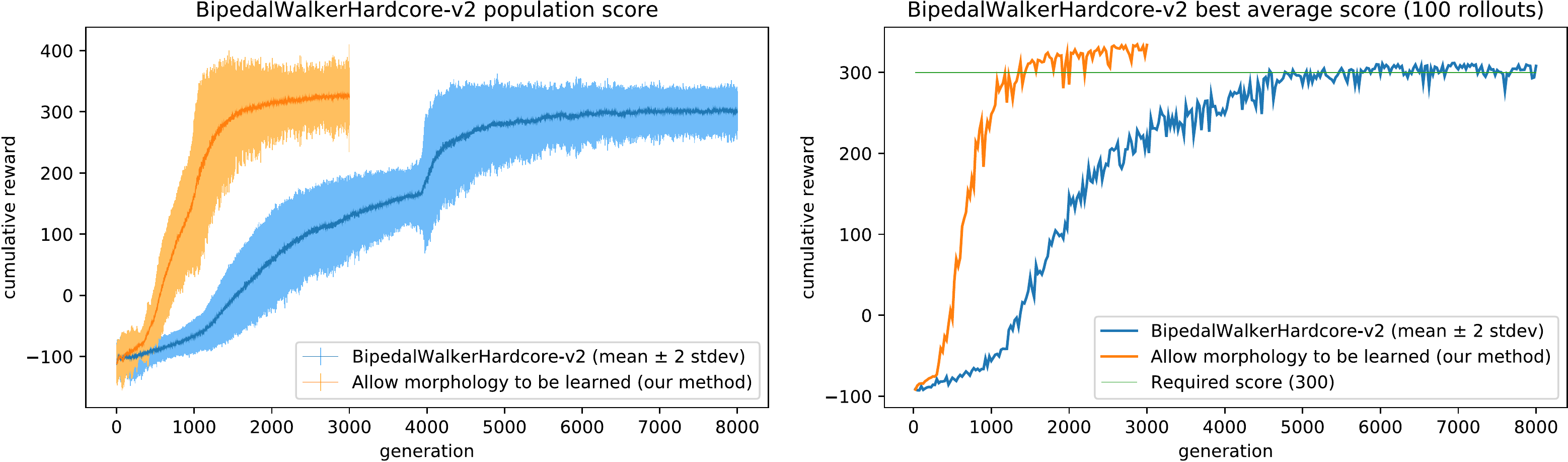}}
\vskip -0.07 in
\caption{Population-based training curves for both versions of \texttt{BipedalWalkerHardcore-v2} (left). Plot of performance of best agent in the population over 100 random trials (right). Original version solved in under 4600 generations (40 hours); By allowing morphology to be learned, the task is solved in under 1400 generations (12 hours).}
\label{fig:bipedhard_learning}
\end{center}
\vskip -0.20 in
\end{figure}

In this environment, our agent generally learns to develop longer, thinner legs, with the exception in the rear leg where it developed a thicker lower limb to serve as useful stability function for navigation. Its front legs, which are smaller and more manoeuvrable, also act as a sensor for dangerous obstacles ahead that complement its LIDAR sensors. While learning to develop this newer structure, it jointly learns a policy to solve the task in 30\% of the time it took the original, static version of the environment. The average scores over 100 rollouts for the learnable version is 335 $\pm$ 37 compared to the baseline score of 313 $\pm$ 53. The full results are summarized in Table~\ref{tab:biped}.

\begin{table}[!htb]
\vskip -0.05in
\begin{center}
\begin{small}
\begin{tabular}{llccccccccc}
\toprule
\multicolumn{3}{l}{\texttt{BipedalWalker-v2}} & \multicolumn{2}{l}{Top leg 1} & \multicolumn{2}{l}{Bottom leg 1} & \multicolumn{2}{l}{Top leg 2} & \multicolumn{2}{l}{Bottom leg 2} \\
 & Avg. score & leg area & $w$ & $h$ & $w$ & $h$ & $w$ & $h$ & $w$ & $h$ \\ 
\midrule
Original         & 347 $\pm$ 0.9 & 100\% & 8.0   & 34.0   & 6.4  & 34.0   & 8.0   & 34.0   & 6.4 & 34.0   \\
Learnable    & 359 $\pm$ 0.2 & 33\%  & 2.0 & 57.3 & 1.6  & 46.0 & 2.0 & 48.8 & 1.6 & 18.9 \\
Reward smaller leg & 323 $\pm$ 68  & 8\%   & 2.0 & 11.5 & 1.6  & 10.6 & 2.0 & 11.4 & 1.6 & 10.2 \\
\midrule
\\
\multicolumn{3}{l}{\texttt{BipedalWalkerHardcore-v2}} & \multicolumn{2}{l}{Top leg 1} & \multicolumn{2}{l}{Bottom leg 1} & \multicolumn{2}{l}{Top leg 2} & \multicolumn{2}{l}{Bottom leg 2} \\
 & Avg. score & leg area & $w$ & $h$ & $w$ & $h$ & $w$ & $h$ & $w$ & $h$ \\ 
\midrule
Original & 313 $\pm$ 53  & 100\% & 8.0   & 34.0   & 6.4  & 34.0   & 8.0   & 34.0   & 6.4 & 34.0  \\
Learnable    & 335 $\pm$ 37  & 95\%  & 2.7 & 59.3 & 10.0 & 58.9 & 2.3 & 55.5 & 1.7 & 34.6 \\
Reward smaller leg & 312 $\pm$ 69  & 27\%  & 2.0 & 35.3 & 1.6  & 47.1 & 2.0 & 36.2 & 1.6 & 26.7 \\
\bottomrule
\end{tabular}
\end{small}
\end{center}
\caption{Summary of results for Bipedal Walker environments. Scaled Box2D dimensions reported.}
\label{tab:biped}
\vskip -0.2 in
\end{table}

\subsection{Optimize for both the task and desired design properties}
\label{section:optimizedesignproperties}

Allowing an agent to learn a better version of its body obviously enables it to achieve better performance. But what if we want to give back some of the additional performance gains, and also optimize also for desirable design properties that might not generally be beneficial for performance? For instance, we may want our agent to learn a design that utilizes the least amount of materials while still achieving satisfactory performance on the task. Here, we reward an agent for developing legs that are smaller in area, and augment its reward signal during training by scaling the rewards by a utility factor of $1+\log(\frac{\text{orig leg area}}{\text{new leg area}})$. Augmenting the reward encourages development of smaller legs.

\begin{figure}[!htb]
\vskip -0.05in
\begin{center}
\centerline{\includegraphics[width=1.0\columnwidth]{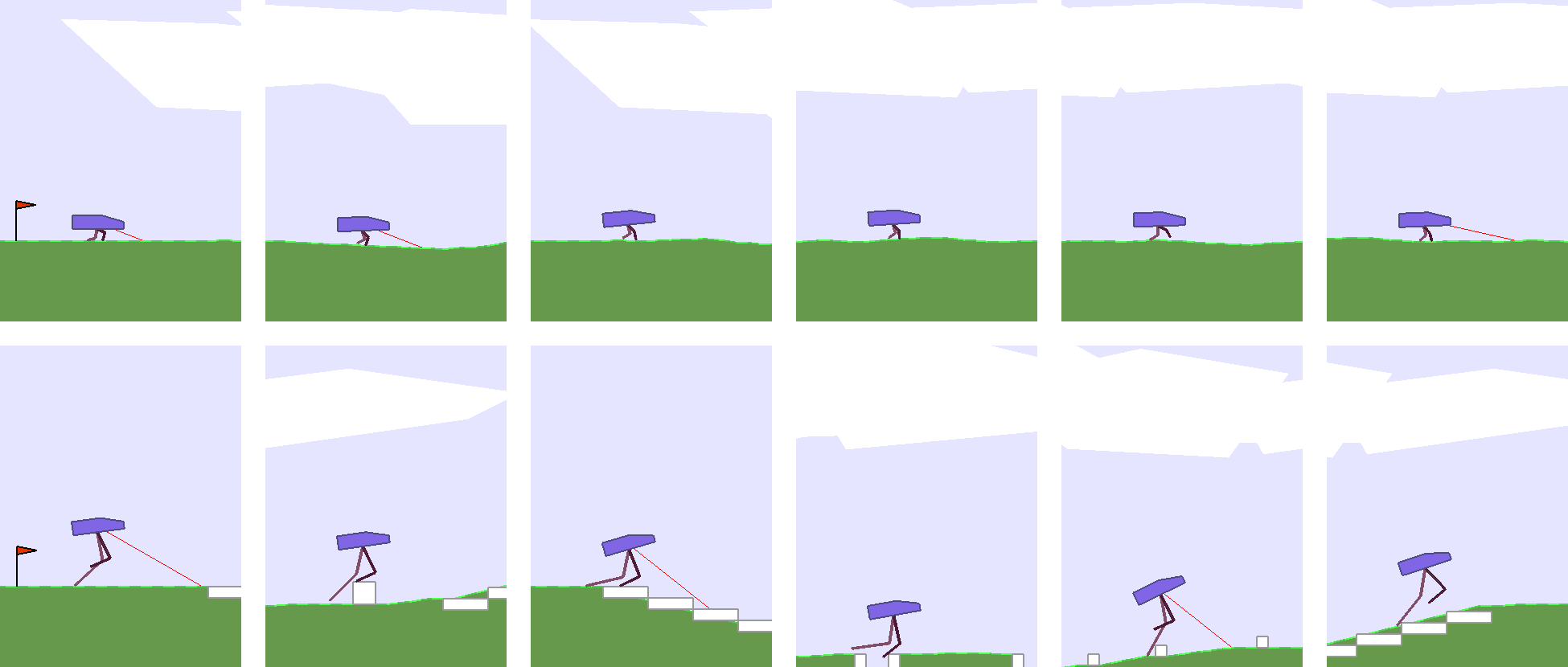}}
\vskip -0.07 in
\caption{Agent rewarded for smaller legs for the task in \texttt{BipedalWalker-v2} environment~\cite{openai_gym} (top). Agent learns the smallest pair of legs that still can solve \texttt{BipedalWalkerHardcore-v2} (bottom).}
\label{fig:biped_small}
\end{center}
\vskip -0.20 in
\end{figure}

This reward augmentation resulted in much a smaller agent that is still able to support the same payload. In \texttt{BipedalWalker}, given the simplicity of the task, the agent's leg dimensions eventually shrink to near the lower bound of $\sim$ 25\% of the original dimensions, with the exception of the heights of the top leg parts which settled at $\sim$ 35\% of the initial design, while still achieving an average (unaugmented) score of 323 $\pm$ 68. For this task, the leg area used is 8\% of the original design.

However, the agent is unable to solve the more difficult \texttt{BipedalWalkerHardcore} task using a similar small body structure, due to the various obstacles presented. Instead, it learns to set the widths of each leg part close to the lower bound, and instead learn the shortest heights of each leg part required to navigate, achieving a score of 312 $\pm$ 69. Here, the leg area used is 27\% of the original.

\subsection{Results over Multiple Experimental Runs}
\label{section:multiple_runs}

In the previous sections, for simplicity, we have presented results over a single representative experimental run to convey qualitative results such as morphology description corresponding to average score achieved. Running the experiment from scratch with a different random seed may generate different morphology designs and different policies that lead to different performance scores. To demonstrate that morphology learning does indeed improve the performance of the agent over multiple experimental runs, we run each experiment 10 times and report the full range of average scores obtained in Table~\ref{tab:biped_multible_runs} and Table~\ref{tab:biped_multible_runs_full_data}. From multiple independent experimental runs, we see that morphology learning consistently produces higher scores over the normal task.

\begin{table}[!htb]
\vskip -0.05in
\begin{center}
\begin{small}
\begin{tabular}{lc}
\toprule
Experiment & Statistics of average scores over 10 independent runs \\ 
\midrule
(a) Ant & 3139 $\pm$ 189.3 \\
(b) Ant + Morphology & 5267 $\pm$ 631.4 \\
\midrule
(c) Biped & 345 $\pm$ 1.3\\
(d) Biped + Morphology & 354 $\pm$ 2.2 \\
(e) Biped + Morphology + Smaller Leg & 330 $\pm$ 3.9 \\
\midrule
(f) Biped Hardcore & 300 $\pm$ 11.9 \\
(g) Biped Hardcore + Morphology & 326 $\pm$ 12.7 \\
(h) Biped Hardcore + Morphology + Smaller Leg & 312 $\pm$ 11.9 \\
\bottomrule
\end{tabular}
\end{small}
\end{center}
\caption{Summary of results for each experiment over 10 independent runs.}
\label{tab:biped_multible_runs}
\vskip -0.2 in
\end{table}

\begin{table}[!htb]
\vskip -0.05in
\begin{center}
\begin{small}
\begin{tabular}{lllllllllll}
\toprule
Experiment & \#1 & \#2 & \#3 & \#4 & \#5 & \#6 & \#7 & \#8 & \#9 & \#10 \\ 
\midrule
(a) Ant & 3447 & 3180 & 3076 & 3255 & 3121 & 3223 & 3130 & 3096 & 3167 & 2693 \\
(b) + Morphology & 5789 & 6035 & 5784 & 4457 & 5179 & 4788 & 4427 & 5253 & 6098 & 4858 \\
\midrule
(c) Biped & 347 & 343 & 347 & 346 & 345 & 345 & 345 & 346 & 346 & 344 \\
(d) + Morphology & 359 & 354 & 353 & 354 & 353 & 352 & 353 & 352 & 353 & 356 \\
(e) + Smaller Leg & 323 & 327 & 327 & 331 & 330 & 331 & 333 & 329 & 337 & 333 \\
\midrule
(f) Biped Hardcore & 313 & 306 & 300 & 283 & 311 & 295 & 307 & 309 & 292 & 279 \\
(g) + Morphology & 335 & 331 & 330 & 330 & 332 & 292 & 327 & 331 & 316 & 330 \\
(h) + Smaller Leg & 312 & 320 & 314 & 318 & 307 & 314 & 316 & 281 & 319 & 324 \\
\bottomrule
\end{tabular}
\end{small}
\end{center}
\caption{Full results from each of the 10 experimental trials. Each number is the average score of the trained agent over 100 rollouts in the environment.}
\label{tab:biped_multible_runs_full_data}
\vskip -0.2 in
\end{table}

\begin{figure}[!htb]
\vskip -0.05in
\begin{center}
\centerline{\includegraphics[width=1.0\columnwidth]{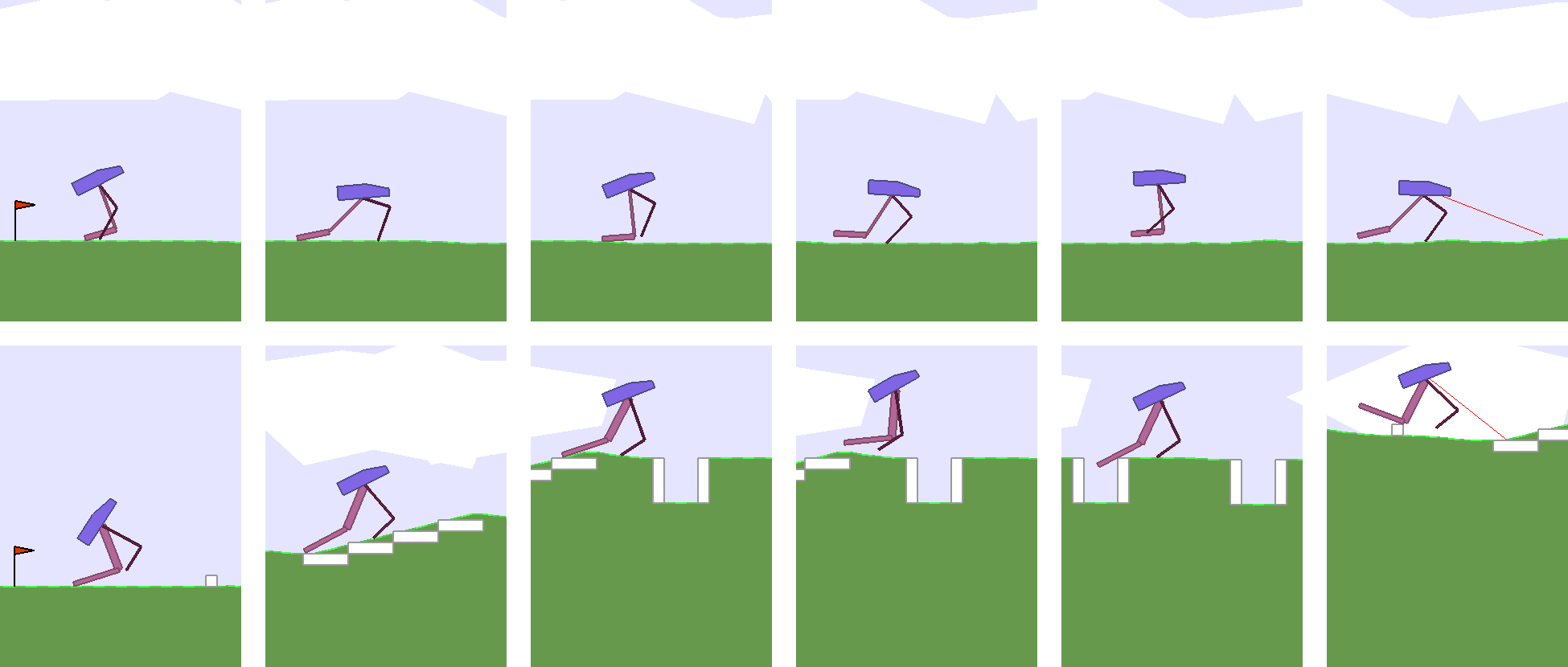}}
\vskip -0.07 in
\caption{Examples of learned morphology in Run \#9. Biped + Morphology (top) develops a thicker but short rear lower limb, unlike the agent in Figure~\ref{fig:bullet_ant}. Biped Hardcore + Morphology (bottom) develops a larger rear leg, but unlike the agent in Figure~\ref{fig:bipedhard}, its thigh is larger than the lower limb.}
\label{fig:multi_run_examples}
\end{center}
\vskip -0.20 in
\end{figure}

We also visualize the variations of morphology designs over different runs in Figure~\ref{fig:multi_run_examples} to get a sense of the variations of morphology that can be discovered during training. As these models may take up to several days to train for a particular experiment on a powerful 96-core CPU machine, it may be costly for the reader to fully reproduce the variation of results here, especially when 10 machines running the same experiment with different random seeds are required. We also include all pretrained models from multiple independent runs in the GitHub repository containing the code to reproduce this paper. The interested reader can examine the variations in more detail using the pretrained models.

\section{Discussion and Future Work}

We have shown that allowing a simple population-based policy gradient method to learn not only the policy, but also a small set of parameters describing the environment, such as its body, offer many benefits. By allowing the agent's body to adapt to its task within some constraints, it can learn policies that are not only better for its task, but also learn them more quickly.

The agent may discover design principles during this process of joint body and policy learning. In both \texttt{RoboschoolAnt} and \texttt{BipedalWalker} experiments, the agent has learned to break symmetry and learn relatively larger rear limbs to facilitate their navigation policies. While also optimizing for material usage for \texttt{BipedalWalker}'s limbs, the agent learns that it can still achieve the desired task even by setting the size of its legs to the minimum allowable size. Meanwhile, for the much more difficult \texttt{BipedalWalkerHardcore-v2} task, the agent learns the appropriate length of its limbs required for the task while still minimizing the material usage.

This approach may lead to useful applications in machine learning-assisted design, in the spirit of~\cite{carter2017,carter2016}. While not directly related to agent design, machine learning-assisted approaches have been used to procedurally generate game environments that can also facilitate policy learning of game playing agents~\cite{togelius2008experiment,millington2009artificial,summerville2018procedural,volz2018evolving,guzdial2018co}. Game designers can optimize the designs of game character assets while at the same time being able to constrain the characters to keep the essence of their original forms. Optimizing character design may complement existing work on machine learning-assisted procedural content generation for game design. By framing the approach within the popular OpenAI Gym framework, design firms can create more realistic environments -- for instance, incorporate strength of materials, safety factors, malfunctioning of components under stressed conditions, and plug existing algorithms into this framework to optimize also for design aspects such as energy usage, easy-of-manufacturing, or durability. The designer may even incorporate aesthetic constraints such as symmetry and aspect ratios that suits her design sense.

In this work we have only explored using a simple population-based policy gradient method~\cite{williams1992} for learning. State-of-the-art model-free RL algorithms, such as TRPO~\cite{schulman2015trust} and PPO~\cite{schulman2017proximal} work well when our agent is presented with a well-designed dense reward signal, while population-based methods offer computational advantages for sparse-reward problems~\cite{openai_es,DeepNeuroevolution2017}. In our setting, as the body design is parameterized by a small set of learnable parameters and is only set once at the beginning of a rollout, the problem of learning the body along with the policy becomes more sparse. In principle, we could allow an agent to augment its body \textit{during} a rollout to obtain a dense reward signal, but we find this unpractical for realistic problems. Future work may look at separating the learning from dense-rewards and sparse-rewards into an inner loop and outer loop, and also examine differences in performance and behaviours in structures learned with various different RL algorithms.

Separation of policy learning and body design into inner loop and outer loop will also enable the incorporation of evolution-based approaches to tackle the vast search space of morphology design, while utilizing efficient RL-based methods for policy learning. The limitations of the current approach is that our RL algorithm can learn to optimize only existing design properties of an agent's body, rather than learn truly novel morphology in the spirit of Karl Sims' \textit{Evolving Virtual Creatures}~\cite{sims1994evolving}.

Nevertheless, our approach of optimizing the specifications of an existing design might be practical for many applications. While a powerful evolutionary algorithm that can also evolve novel morphology might come up with robot morphology that easily outperforms the best bipedal walkers in this work, the resulting designs may not be as useful to a game designer who is tasked to work explicitly with bipedal walkers that fit within the game's narrative (although it is debatable whether the game can be more entertaining and interesting if the designer is allowed to explore the space beyond given specifications). Due to the vast search space of all possible morphology, a search algorithm can easily come up with unrealistic or unusable designs that exploits its simulation environment, as discussed in detail in \cite{lehman2018surprising}, which may be why subsequent morphology evolution approaches constrain the search space of the agent's morphology, such as constraining to the space of soft-body voxels~\cite{cheney2013unshackling} or constraining to a set of possible pipe frame connection settings~\cite{jansen2008strandbeests}. We note that unrealistic designs may also result in our approach, if we do not constrain the learned dimensions to be within $\pm$ 75\% of its original value. For some interesting examples of what REINFORCE discovers without any constraints, we invite the reader to view the \textit{Bloopers} section of \url{https://designrl.github.io/}. 

Just as REINFORCE~\cite{williams1992} can also be applied to the discrete search problem of neural network architecture designs~\cite{zoph2016neural}, similar RL-based approaches could be used for novel morphology design -- not simply for improving an existing design like in this work. We believe the ability to learn useful morphology is an important area for the advancement of AI. Although morphology learning originally initiated from the field of evolutionary computation, we hope this work will engage the RL community to investigate the concept further and encourage idea exchange across communities.

\section*{Acknowledgments}

We would like to thank the three reviewers from Artificial Life Journal, Luke Metz, Douglas Eck, Janelle Shane, Julian Togelius, Jeff Clune, and Kenneth Stanley for their thoughtful feedback and conversation. All experiments were performed on CPU machines provided by Google Cloud.

\newpage

\small

\bibliography{main}
\bibliographystyle{abbrv}

\end{document}